\documentclass[manuscript]{acmart}
\usepackage{threeparttable}
\usepackage{array}
\usepackage{multirow}
\usepackage{color}
\usepackage{tabu}

\AtBeginDocument{%
  \providecommand\BibTeX{{%
    \normalfont B\kern-0.5em{\scshape i\kern-0.25em b}\kern-0.8em\TeX}}}

\begin{document}

\title{RISAM: Referring Image Segmentation via Mutual-Aware Attention Features}

\author{Mengxi Zhang}
\email{mengxizhang@tju.edu.cn}
\orcid{0000-0002-6011-1218}
\affiliation{%
  \institution{Tianjin University}
  \streetaddress{Weijin Road, No. 92.}
  \city{Tianjin}
  \country{China}
  \postcode{300072}
}

\author{Yiming Liu}
\email{liuyiming@xiaoyingai.com}
\affiliation{%
  \institution{Xiaoying AI lab}
  \streetaddress{}
  \city{Beijing}
  \country{China}}

\author{Xiangjun Yin}
\email{yinxiangjun@tju.edu.cn}
\orcid{0000-0002-4829-9019}
\affiliation{%
  \institution{Tianjin University}
  \streetaddress{Weijin Road, No. 92.}
  \city{Tianjin}
  \country{China}
  \postcode{300072}
}

\author{Huanjing Yue}
\email{huanjing.yue@tju.edu.cn}
\orcid{0000-0003-2517-9783}
\affiliation{%
  \institution{Tianjin University}
  \streetaddress{Weijin Road, No. 92.}
  \city{Tianjin}
  \country{China}
  \postcode{300072}
}

\author{Jingyu Yang}
\email{yjy@tju.edu.cn}
\orcid{0000-0002-7521-7920}
\affiliation{%
  \institution{Tianjin University}
  \streetaddress{Weijin Road, No. 92.}
  \city{Tianjin}
  \country{China}
  \postcode{300072}
}


\begin{abstract}
Referring image segmentation (RIS) aims to segment a particular region based on a language expression prompt. Existing methods incorporate linguistic features into visual features and obtain multi-modal features for mask decoding. However, these methods may segment the visually salient entity instead of the correct referring region, as the multi-modal features are dominated by the abundant visual context. In this paper, we propose RISAM, a referring image segmentation method that leverages the Segment Anything Model (SAM) based on the parameter-efficient fine-tuning framework and introduces the mutual-aware attention mechanism to get the accurate referring mask. 
Specifically, our mutual-aware attention mechanism consists of Vision-Guided Attention and Language-Guided Attention, which bidirectionally model the relationship between visual and linguistic features. Correspondingly, we design the Mutual-Aware Mask Decoder to enable extra linguistic guidance for more consistent segmentation with the language expression. To this end, a multi-modal query token is introduced to integrate linguistic information and interact with visual information simultaneously. Extensive experiments on three benchmark datasets, $i.e.$, RefCOCO, RefCOCO+, and G-Ref show that our method outperforms the state-of-the-art RIS methods. Besides, our model also shows great performance in terms of generalization ability and multi-object segmentation on PhraseCut and gRefCOCO, respectively. Our code will be available at \href{https://github.com/TakoCC/RISAM/}{https://github.com/TakoCC/RISAM}.
\end{abstract}


%

\maketitle

\section{Introduction}

 \begin{figure}[t]
	\centering
        \includegraphics[scale=0.23]{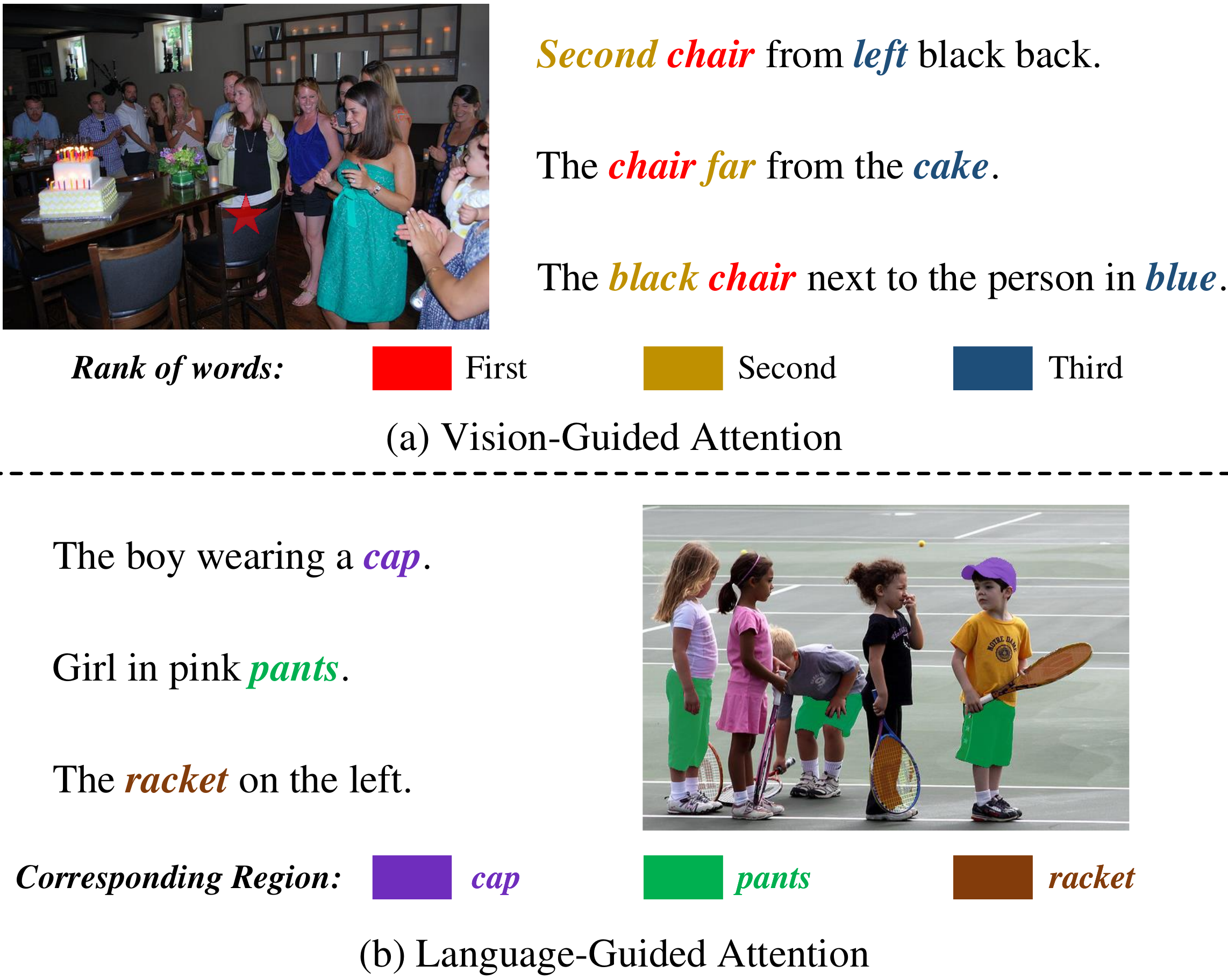}
	\caption{The illustration of Vision-Guided Attention (a) and Language-Guided Attention (b). For Vision-Guided Attention, we list the three most informative words for the image region symbolized by a red pentangle. For Language-Guided Attention, the most corresponding regions for each word, $e.g.$, `cap', `pants', and `racket', are denoted by different colors. Previous RIS methods only consider Vision-Guided Attention to fuse visual and linguistic features, but none of these methods introduce Language-Guided Attention to generate vision-aware linguistic features and further use them in the mask decoder.}
    \label{framework}
\end{figure}

Referring image segmentation~\cite{hu2016segmentation} (RIS) is a fundamental and challenging multi-modal task, which involves both vision-language understanding~\cite{glip,csvt9} and instance segmentation~\cite{maskrcnn, csvt10}. The target of RIS is to locate particular regions according to the given query in natural language. It has great potential in many applications, $e.g.$, human-machine interaction and interactive image segmentation. 

Existing RIS methods~\cite{csvt1, csvt2, BRINET,vlt,gres} introduce various fusion methods to obtain multi-modal features. Then, these features are sent into a mask decoder to predict the segmentation mask. Despite significant advancements in RIS, there are still several limitations. First, current methods~\cite{vlt, gres} utilize the unidirectional attention mechanism to fuse features from different modalities. However, they only consider the linguistic guidance for visual features but ignore the visual guidance for linguistic features. Unlike the unidirectional attention mechanism, BRINet~\cite{BRINET} adopts both visual and linguistic guidance in a serial bidirectional way. Nevertheless, due to the serial manner, it only generates vision-aware linguistic features in the fusion model but does not further use language-aware visual features in the mask decoder. Second, existing methods send the multi-modal features into a mask decoder to generate the final segmentation mask. However, since multi-modal features are produced by integrating linguistic properties into visual features, they are still dominated by visual properties. Without extra linguistic guidance, the mask decoder will focus on the most visually salient entities but ignore linguistic consistency. Moreover, existing methods typically fine-tune the encoders to adapt them for the datasets of RIS. However, this strategy shrinks the generalization ability of encoders pre-trained on a large-scale dataset.


In this paper, we propose a novel method that utilizes the mutual-aware attention mechanism and transfers the knowledge of Segment Anything Model (SAM)~\cite{sam} into RIS by the parameter-efficient fine-tuning (PEFT) framework. 
First, we introduce the Mutual-Aware Attention block to bidirectionally model the relationship between visual and linguistic features. The Mutual-Aware Attention block consists of two parallel branches: Vision-Guided Attention and Language-Guided Attention. As shown in Fig.~\ref{framework}, Vision-Guided Attention assigns different weights to each word in the expression for each image region (such as the red pentangle) and produces language-aware visual features. Similarly, Language-Guided Attention explores the corresponding image region for the word, $e.g.$, \emph{`cap'}, \emph{`pants'}, \emph{`racket'}, and generates vision-aware linguistic features. We consider language-aware visual features and vision-aware linguistic features as the mutual-aware attention features of our method. 
Second, we design the Mutual-Aware Mask Decoder to enable extra linguistic guidance. In particular, we introduce a multi-modal query token to integrate visual and linguistic properties, which helps to segment the correct referring region.
Finally, we introduce the parameter-efficient fine-tuning (PEFT) framework to preserve the generalization ability of pre-trained encoders. To transfer the knowledge of SAM into RIS, we introduce a Feature Enhancement module to integrate global and local visual features.

\begin{figure*}[t]
	\centering
        \includegraphics[scale=0.14]{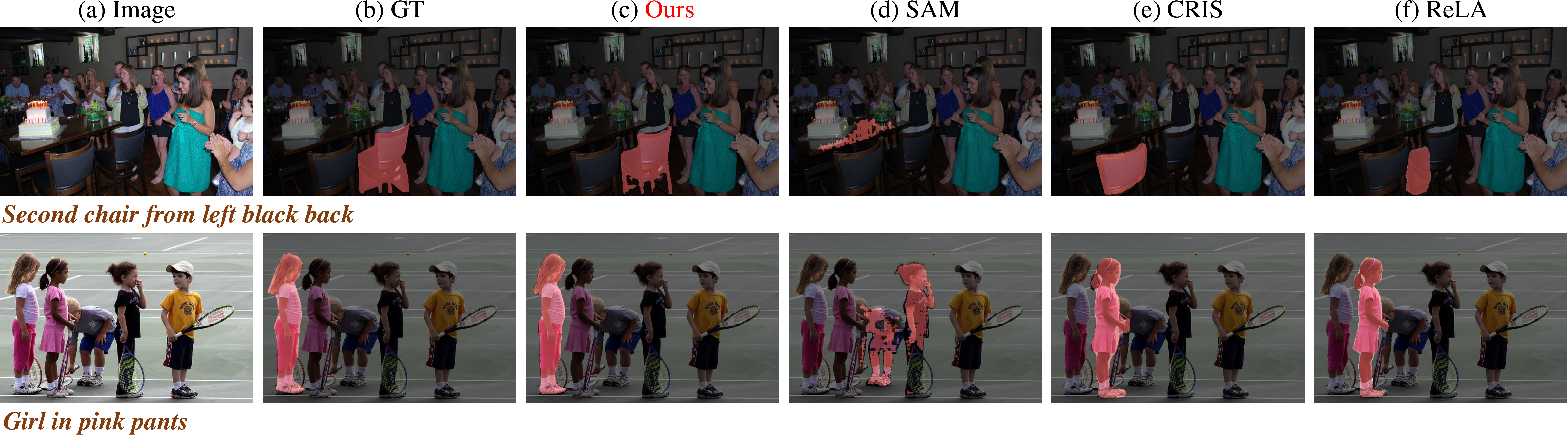}
	\caption{ Segmentation masks generated by our method (c) and other methods, including directly using SAM~\cite{sam} (d), CRIS~\cite{cris} (e), and ReLA~\cite{gres} (f). Directly using SAM means training the SAM decoder only. }
        \label{intro2}
\end{figure*}

We demonstrate the results of our method and other methods in Fig.~\ref{intro2}. To our knowledge, our work is the first to transfer the powerful knowledge of SAM into RIS. To be summarized, our contributions are listed as follows:

$\bullet$ We propose a referring image segmentation method called RISAM, which leverages the powerful knowledge of SAM based on the parameter-efficient fine-tuning framework and uses the mutual-aware attention features to get a language-consistent mask. 

$\bullet$ We introduce a Mutual-Aware Attention block to produce language-aware visual features and vision-aware linguistic features by weighting each word of the sentence and each region of visual features. Besides, we design the Mutual-Aware Mask Decoder to utilize extra linguistic guidance and get a segmentation mask consistent with the language expression. In particular, we introduce a multi-modal query token to integrate visual and linguistic properties.

$\bullet$ The proposed approach achieves new state-of-the-art performance on the three widely used RIS datasets, including RefCOCO, RefCOCO+, and G-Ref. Additionally, our method exhibits excellent capabilities in terms of generalization and multi-object segmentation ability. 
\begin{figure*}[t]
	\centering{\includegraphics[scale=0.24]{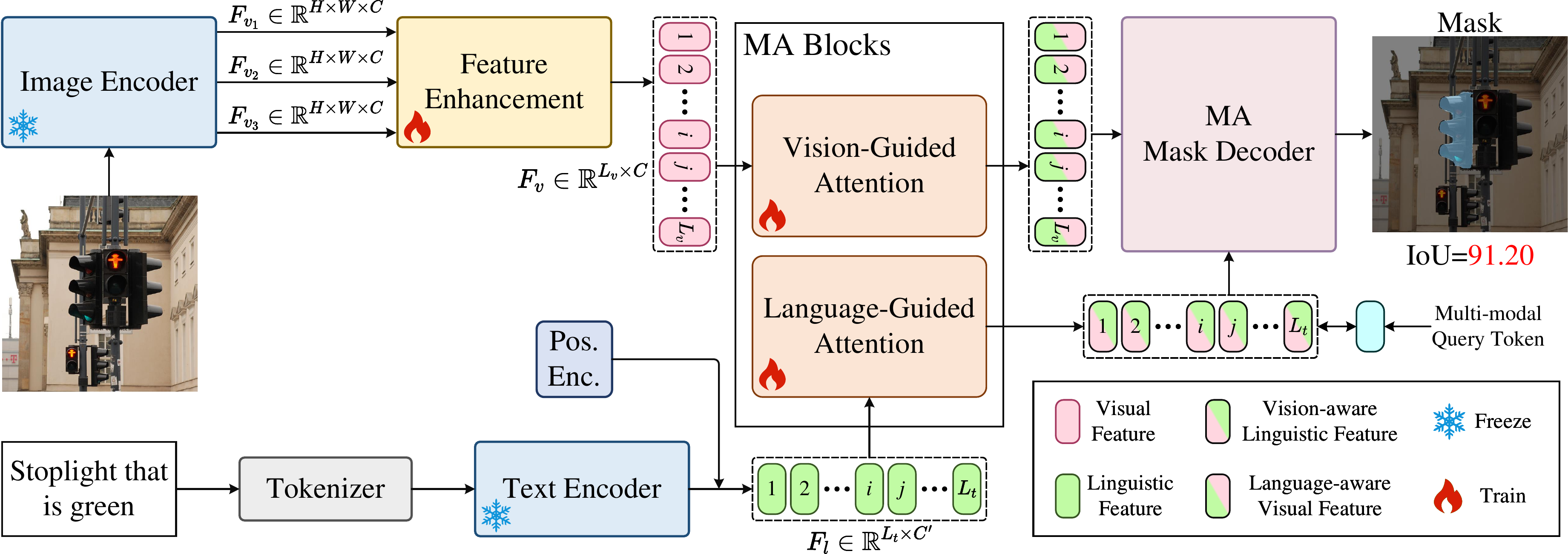}}
	\caption{ The overview of RISAM. For an input image, the image encoder extracts shallow/middle/deep visual features ($F_{v_1},F_{v_2},F_{v_3}$). For the language expression, the text encoder generates linguistic features ($F_l$). Then, these features are sent into the Feature Enhancement module and obtain enhanced visual features ($F_{v}$). Subsequently, Mutual-Aware Attention (MA) blocks receive enhanced visual features and linguistic features as inputs to get mutual-aware attention features. After that, the Mutual-Aware Mask Decoder utilizes a multi-modal query token and mutual-aware attention features to get the final segmentation mask. (Pos. Enc. symbolizes position encodings of linguistic features.)}
 \label{overview}
\end{figure*}
\section{Related Work}

\subsection{Referring Image Segmentation}
Referring image segmentation~\cite{hu2016segmentation} aims to segment a particular region according to the natural language expression. Early approaches~\cite{yu2016modeling,liu2017recurrent,yu2018mattnet,liu2019learning,luo2020multi}  concatenate visual and linguistic features to produce multi-modal features, which are fed into the fully convolutional network for segmentation generalization. Yu et al.~\cite{yu2018mattnet} proposed a two-stage method that first generates masks by Mask R-CNN~\cite{maskrcnn}, and then selects the target mask with the linguistic prompt. BRINet~\cite{BRINET} builds a serial bi-directional relationship between the language and image. Besides, MCN~\cite{luo2020multi} presented a multi-task framework to jointly optimize two related tasks, \emph{i.e.}, referring expression comprehension and segmentation.

As the attention mechanism~\cite{attentions,nonlocal} achieved great success in various fields, it has been exploited in the field of RIS~\cite{shi2018key,cmsa,BRINET,csvt1,csvt2}. Shang et al.~\cite{csvt1} design a recurrent network structure to obtain a more comprehensive global semantic understanding. Later, some methods~\cite{lavt,kim2022restr,vlt} adopt the transformer-based architectures. VLT~\cite{vlt} introduces a Vision-Language Transformer to enhance deep interactions among multi-modal features. More recently, CRIS~\cite{cris} utilized CLIP~\cite{clip} as the image and text encoder and transferred the knowledge of CLIP for text-to-pixel alignment. CGFormer~\cite{cgformer} replaces the conventional pixel classification framework with the mask classification framework for RIS. ReLA~\cite{gres} introduces a new task called generalized referring expression segmentation, which enables expressions to indicate the existence of target objects. However, these methods fail to produce and further utilize vision-aware linguistic features in the mask decoder.

\subsection{Attention Mechanism in Multi-modal Tasks}
Attention mechanism has been widely used in various multi-modal tasks. In~\cite{unimo,VLP,csvt4,csvt6}, Transformer schemes are used to exploit the long-range dependencies between visual features and linguistic features. Besides, the Transformer-decoder based architectures~\cite{albef,blipv2} are also used to fuse the visual and linguistic features. For example, BLIP-2~\cite{blipv2} builds a Q-former based on the cross-attention mechanism to break the gap between the visual and linguistic modalities by a set of learned queries. Wen et al.~\cite{csvt5} use the graph attention mechanism to capture both object-level visual relations and global-regional visual relations for image-text matching. Recently, Multi-modal Large Language Models~\cite{llava, kosmos} achieve fantastic performance on various vision-language understanding tasks. A key component for this success is that the self-attention mechanism captures the complicated relationships among different modalities effectively. Existing RIS methods only use features from the vision branch for subsequent mask decoding, which leads the mask decoder to segment the visually salient entities. In this paper, we propose a Mutual-Aware Attention scheme to generate language-aware visual features and vision-aware linguistic features, where the latter acts as the extra linguistic guidance to generate an accurate referring mask.


\subsection{Powerful Foundation Models in Computer Vision}


Foundation models are trained on broad data and can be adapted ($e.g.$, fine-tuned) to a wide range of downstream tasks. In recent years, some vision transformers~\cite{vit,swin} achieved state-of-the-art performance on various tasks, including image classification, semantic/instance segmentation, and object detection. Due to great efforts made on large-scale datasets, recent foundation models~\cite{clip,sam} are equipped with more powerful feature representations. Benefiting from 400 million image-text pairs, CLIP~\cite{clip} achieved strong zero-shot ability on many visual tasks. Some researchers utilize the knowledge of CLIP for different tasks, including semantic segmentation~\cite{groupvit}, object detection~\cite{vild,glip}, and referring image segmentation~\cite{cris}. Recently, Meta released SAM, the first segmentation foundation model trained on more than 1 billion masks, and achieved remarkable performance on interactive segmentation. 
In this paper, we first adopt the PEFT framework to transfer the powerful knowledge of SAM into the field of RIS. Besides, we design the Mutual-Aware Attention block and the Mutual-Aware Mask Decoder to get a high-quality segmentation mask.

\section{Method}
The overall architecture of RISAM is shown in Fig.~\ref{overview}. Firstly, the input image and language expression are projected into the visual ($F_{v_1}$, $F_{v_2}$, $F_{v_3}$) and linguistic ($F_l$) feature spaces via a pre-trained image encoder~\cite{sam} and a text encoder~\cite{clip}, respectively. Secondly, we design a Feature Enhancement (FE) module, which fuses features from different layers of the image encoder and obtains enhanced visual features ($F_{v}$). Thirdly, enhanced visual features ($F_{v}$) and linguistic features ($F_{l}$) are fed into the Mutual-Aware Attention (MA) blocks to obtain mutual-aware attention features, including vision-aware linguistic features and language-aware visual features. Finally, we design the Mutual-Aware Mask Decoder to use mutual-aware attention features effectively and produce a language-consistent mask. Besides, to preserve the generalization ability of the image and text encoder and save computational costs, we design a parameter-efficient fine-tuning framework for referring image segmentation. Specifically, we set the parameters of encoders frozen and only update other modules, which occupy around $1\%$ parameters of the whole model. We will describe the details of these steps in the following subsections.

\subsection{Image Encoder and Text Encoder}
The image encoder of SAM~\cite{sam}, a VIT-based backbone, takes images of size $4H\times4W$ as inputs and generates visual features of spatial size $H\times W$. In particular, SAM uses a VIT-H with $14\times14$ windows and four plain global attention blocks. For an input image, we utilize visual features from 2nd$\sim$4th global attention blocks, which are defined as shallow layer features $F_{v_1}\in \mathbb{R}^{H \times W \times C}$, middle layer features $F_{v_2}\in \mathbb{R}^{H \times W \times C}$ and deep layer features $F_{v_3}\in \mathbb{R}^{H \times W \times C}$. Here, $H$ and $W$ are the height and width of the feature map, respectively, and $C$ denotes the channel size of visual features.

For the language expression, we adopt a text encoder pre-trained by~\cite{clip} and obtain linguistic features $F_{l} \in \mathbb{R}^{L_t\times C'}$. Here, $L_t$ denotes the length of linguistic features. Accordingly, $C'$ is the channel size of linguistic features.

To preserve the generalization capability of the image and text encoder and save computational resources, we freeze the parameters of these encoders, which also prevents catastrophic forgetting~\cite{toneva2018empirical}.

\subsection{Feature Enhancement}
A typical RIS language prompt may contain the overall depiction or some detailed descriptions of a specific object. For example, \emph{`Man in the left'} refers to the man by the overall depiction {`left'}; while \emph{`A smiling girl with gloves'} describes the girl with some detailed descriptions, $i.e.$, \emph{`smiling'} and \emph{`with gloves'}. Therefore, to generate an accurate segmentation mask for the causal language expression, it is necessary to focus on both global/semantic information and local/grained details. For the image encoder of SAM, features from the deep layer and shallow layer contain accurate semantic features and abundant grained features, respectively. Based on this consideration, we introduce the Feature Enhancement module to get enhanced visual features, which contain both semantic and grained information.

First, we fuse the shallow layer feature $F_{v_1}$ and the middle layer feature $F_{v_2}$ as follows.
\begin{equation}
\hat{F} = \mathrm{CBA}([\mathrm{MLP}(F_{v_1}), \mathrm{MLP}(F_{v_2})]),
\label{FM:1}
\end{equation}
where $\hat{F}$ denotes the early enhanced feature. $\mathrm{CBA}(\cdot)$ is sequential operations, including convolution layers with $3\times 3$ kernels, a batch-normalization layer, and $GeLu$ activation function. $\mathrm{MLP}(\cdot)$ represents the Multi-Layer Perceptron (MLP) layer. $[\cdot, \cdot]$ is the concatenation operation.

Subsequently, we fuse the early enhanced feature $\hat{F}$ and the deep layer feature $F_{v_3}$ to obtain the final enhanced visual feature, 
\begin{equation}
F_{v} = \mathrm{CBA}([\mathrm{MLP}(\hat{F}), \mathrm{MLP}(F_{v_3})]),
\label{FM:2}
\end{equation}
where $F_{v}\in\mathbb{R}^{H\times W \times C}$ is the final enhanced visual feature. Then the feature map is flattened into a 2-D vector $F_{v}\in\mathbb{R}^{L_v \times C}$, where $L_v$ is equal to $H\times W$.

\subsection{Mutual-Aware Attention Blocks}
\begin{figure}[t]
	\centering{\includegraphics[scale=0.3]{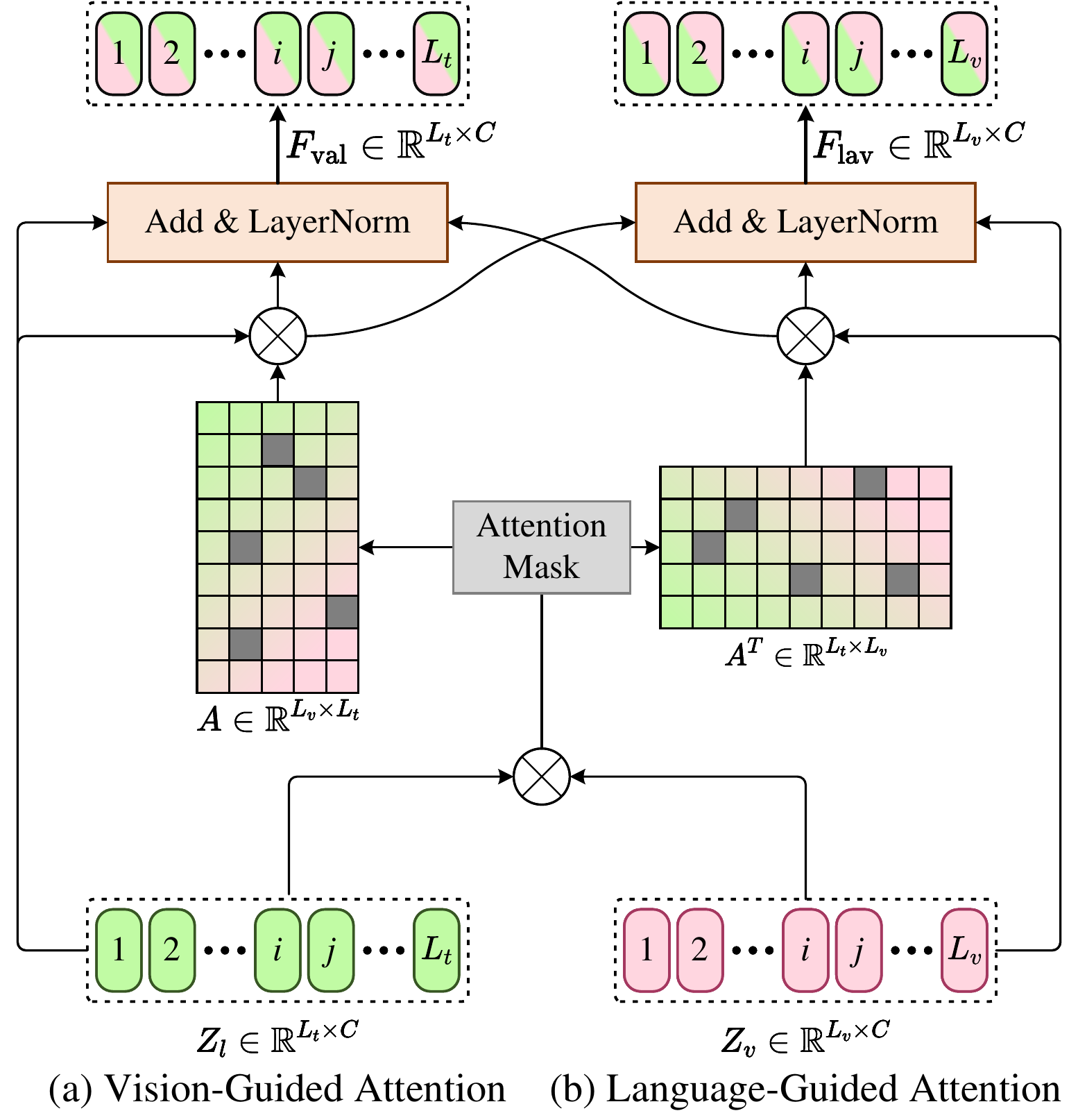}}
	\caption{ The architecture of Mutual-Aware Attention block. The left part (a) is the Vision-Guided Attention branch, and the right part (b) is the Language-Guided Attention branch. $A$ and $A^{\top}$ denote the attention weights. $\otimes$ symbolizes the matmul product operation.}
 \label{pic:VLA}
\end{figure}

After obtaining visual and linguistic features, the first step is to fuse these features and get multi-modal features. Existing methods~\cite{kim2022restr,vlt} propose different strategies to get multi-modal features. However, these methods only assign different weights to each word in the expression but treat each image region equally. BRINet~\cite{BRINET} adopts a serial bidirectional design to utilize both visual and linguistic guidance. However, the serial design fails to further utilize vision-aware linguistic features in the mask decoder. 
To address these issues, we propose the Mutual-Aware Attention block, which consists of two parallel branches. Specifically, the first branch is Vision-Guided Attention, which weights different words for each pixel of visual features. Accordingly, the second branch is Language-Guided Attention, which weights different image regions for each word of the sentence. Besides, we observe that some irrelevant image and text pairs, $e.g.$, assigning the adverb with corresponding image regions, hinder the convergence of Mutual-Aware Attention blocks. Therefore, we introduce the masked attention mechanism into our Mutual-Aware Attention to filter out irrelevant image and text pairs, further improving the performance (reported in Tab.~\ref{tab:abl_vla}). The architecture of Mutual-Aware Attention block is shown in Fig.~\ref{pic:VLA}.

First, we model the correlation between linguistic features and visual features as follows,
\begin{equation}
    \begin{split}
         &Z_v = F_{v}W_{v}, Z_l = F_{l}W_{l}, \\
         &A = \mathrm{Softmax}(\frac{Z_v Z^{\top}_l}{\sqrt{C}} + \mathcal{M}),
    \end{split}
\label{VLA:1}
\end{equation}
where $A\in \mathbb{R}^{L_v\times L_t}$ is the attention weight. $W_{v}$ and $W_{l}$ are learnable matrices of size $C\times C$ and $C'\times C$, which aim to transform $F_{v}$ and $F_{l}$ into the same feature dimension. $\mathcal{M}$ is the attention mask, which is calculated by,

\begin{equation}
         \mathcal{M} (i,j)=
        \left\{\begin{matrix}
        0\quad  &\mathrm{if}M(i,j)<\tau
         \\ -\infty \quad  &\mathrm{otherwise}
        \end{matrix}\right.,
\label{VLA:2}
\end{equation}
where $M=1/(1+e^\frac{Z_v Z^{\top}_l}{\sqrt{C} })$ denotes the relevant scores between visual and linguistic features. $\tau$ is the threshold, and its value will be discussed in the ablation study. Through the attention mask $\mathcal{M}$, we alleviate the interference from irrelevant pairs in visual and linguistic features.

\begin{figure}[t]
	\centering{\includegraphics[scale=0.25]{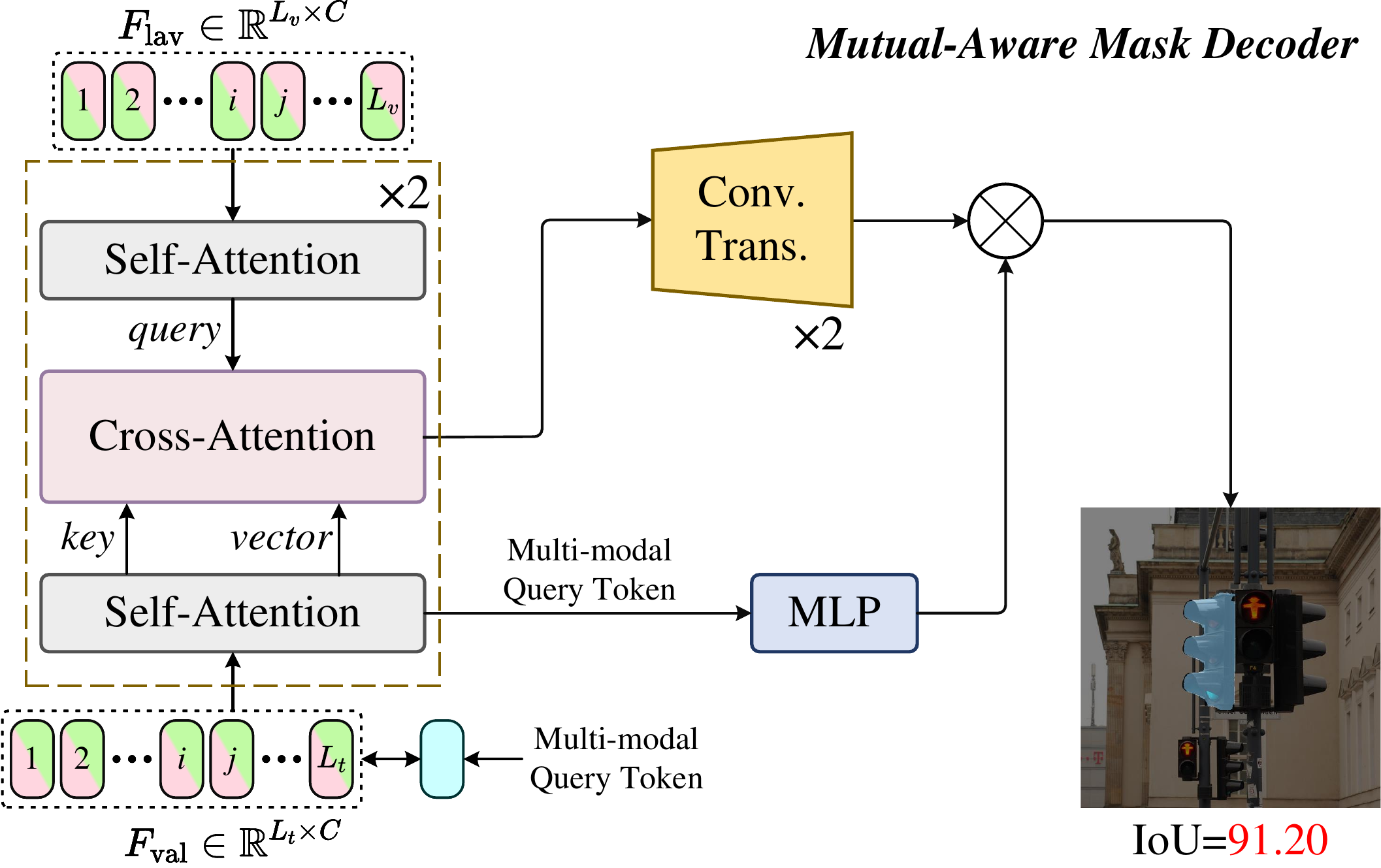}}
	\caption{The architecture of Mutual-Aware Mask Decoder. This mask decoder receives vision-aware linguistic features and language-aware visual features as inputs, where the former acts as the extra linguistic guidance. Additionally, we introduce a multi-modal query token to aggregate linguistic information and interact with visual features, which is beneficial to a high-quality segmentation mask.}
 \label{pic:ELP}
\end{figure}

After that, we obtain mutual-aware attention features, including language-aware visual features $F_{\mathrm{lav}}\in \mathbb{R}^{L_v\times C}$ and vision-aware linguistic features $F_\mathrm{val}\in \mathbb{R}^{L_t\times C}$ as follows,
\begin{equation}
    \begin{split}
         F_{\mathrm{lav}} = &\mathrm{LayerNorm}(AZ_{l} + Z_{v}),  \\
         F_{\mathrm{val}} = &\mathrm{LayerNorm}(A^{\top}Z_{v} + Z_{l}),
    \end{split}
\label{VLA:3}
\end{equation}
where $\mathrm{LayerNorm}(\cdot)$ denotes the Layer Normalization.
 
We use two sequential Mutual-Aware Attention blocks in our implementation, and the number of Mutual-Aware Attention blocks will be discussed in the ablation study.

\subsection{Mutual-Aware Mask Decoder}
After we get mutual-aware attention features, we send them into the mask decoder to obtain the final mask. Previous methods~\cite{cris,vlt,cgformer} only leverage the features from Vision-Guided Attention for subsequent decoding. However, these features still contain excessive visual properties, and the mask decoder is likely to segment visual-dominant entities without extra linguistic guidance. To address this issue, we build the Mutual-aware Mask Decoder based on the mask classification framework~\cite{detr, mask2former}. This module receives vision-aware linguistic feature $F_{val}$ and language-aware visual feature $F_{lav}$ as inputs, where the former contains more linguistic properties and serves as extra linguistic guidance. Besides, we introduce a single multi-modal query token with random initialization. Different from DETR~\cite{detr}/Mask2former~\cite{mask2former}, we combine the multi-modal query token with vision-aware linguistic features as the inputs of the decoder. Such a design enables the multi-modal query token to integrate linguistic information and interact with visual features, thus getting a consistent segmentation with the language expression. The architecture of Mutual-aware Mask Decoder is depicted in Fig.~\ref{pic:ELP}.

To this end, language-aware visual feature $F_{\mathrm{lav}}$ is first fed into a multi-head self-attention layer to extract powerful contextual information.
\begin{equation}
    \hat{F}_{\mathrm{lav}} = \mathrm{MHSA}(F_{\mathrm{lav}}) + F_{\mathrm{lav}},
\label{MD:1}
\end{equation}
where $\mathrm{MHSA}(\cdot)$ is the multi-head self-attention layer, which is composed of three point-wise linear layers mapping $F_{\mathrm{lav}}$ into intermediate representations, $i.e.$, $Q,K,V$, and then calculated as below,
\begin{equation}
    \mathrm{MHSA}(Q,K,V) = \mathrm{Softmax}(\frac{QK^{\top}}{\sqrt{d_k}})V,
    \label{MD:MHSA}
\end{equation}
where $\sqrt{d_k}$ is the feature dimension of $K$. 

Then, the multi-modal query token $F_m \in \mathbb{R}^{1\times C}$ along with  $F_{\mathrm{val}} \in \mathbb{R}^{L_t\times C}$ are sent to a multi-head self-attention layer to aggregate the vision-aware linguistic feature.  
Let $F_c := [F_m, F_{\mathrm{val}}]$, the aggregation is formulated as
\begin{equation}
     \hat{F}_c := [\hat{F}_m, \hat{F}_{\mathrm{val}}] = \mathrm{MHSA}(F_c) + F_c,
\label{MD:3}
\end{equation}
where $\hat{F}_c$ denotes the evolved feature concatenating the evolved versions of vision-aware linguistic feature $\hat{F}_{\mathrm{val}}$ and multi-modal query token $\hat{F}_m$.

Subsequently, we perform interaction between $\hat{F}_{\mathrm{lav}}$ and $\hat{F}_c$, obtaining the evolved language-aware visual feature $\overline{F}_{\mathrm{lav}}$ via a multi-head cross-attention layer as follows.
\begin{equation}
     \overline{F}_{\mathrm{lav}} = \mathrm{MHCA}(\hat{F}_{\mathrm{lav}}, \hat{F}_c, \hat{F}_c) + \hat{F}_{\mathrm{lav}},
\label{MD:4}
\end{equation}
where $\mathrm{MHCA}(\cdot)$ is the multi-head cross-attention layer, which also contains three linear layers to map $\overline{F}_{\mathrm{lav}}$ into $Q$, $\hat{F}_c$ into $K$ and $V$. The following calculation is the same as Eq.~\ref{MD:MHSA}.

\setlength{\tabcolsep}{1pt}
\begin{table*}[t]
\caption{Comparisons with previous state-of-the-art methods in terms of IoU. U: UMD split; G: Google split. $*$, $\dag$, $\ddag$ denote the visual encoders from Mask-RCNN, DeepLab, and CLIP, respectively.}
\label{tab:comp}
\centering
\begin{tabu}{c|c|c|c|ccc|ccc|ccc}
\tabucline[1.2pt]{-}
\multirow{2}{*}{Methods}       & \multirow{2}{*}{\begin{tabular}[c]{@{}c@{}}Vis. \\ Encoder\end{tabular}} & \multirow{2}{*}{\begin{tabular}[c]{@{}c@{}}Text\\ Encoder\end{tabular}} & \multirow{2}{*}{\begin{tabular}[c]{@{}c@{}}Train \\ Params.\end{tabular}} & \multicolumn{3}{c|}{RefCOCO}                     & \multicolumn{3}{c|}{RefCOCO+}                    & \multicolumn{3}{c}{G-Ref}                        \\ \cline{5-13} 
                               &                                                                             &                                                                          &                                                                           & val            & testA          & testB          & val            & testA          & testB          & val (U)        & test (U)       & val (G)        \\ \hline \hline
MAttNet\cite{yu2018mattnet} (CVPR'18)              & Res-101$^*$                                                                     & LSTM                                                                     & 28M                                                                       & 56.51          & 62.37          & 51.70          & 46.67          & 52.39          & 40.08          & 47.64          & 48.61          & -              \\
BRINet\cite{BRINET} (CVPR'20)               & Res-101$^\dag$                                                                      & LSTM                                                                     & 61M                                                                       & 60.98          & 62.99          & 59.21          & 48.17          & 52.32          & 42.11          & -              & -              & 48.04          \\
LTS\cite{lts} (ICCV'21)                  & DarkNet-53                                                                  & GRU                                                                      & 41M                                                                       & 65.43          & 67.76          & 63.08          & 54.21          & 58.32          & 48.02          & 54.40          & 54.25          & -              \\
ReSTR\cite{kim2022restr} (CVPR'22)                & VIT-B                                                                       & Transf.                                                                  & 86M                                                                       & 67.22          & 69.30          & 64.45          & 55.78          & 60.44          & 48.27          & -              & -              & 54.48          \\
CRIS\cite{cris} (CVPR'22)                 & Res-50$^\ddag$                                                                       & CLIP                                                                     & 57M                                                                       & 70.47          & 73.18          & 66.10          & 62.27          & 68.08          & 53.68          & 59.87          & 60.36          & -              \\
RefTR\cite{reftr} (NIPS'21)                & Res-101$^\dag$                                                                      & BERT                                                                     & 61M                                                                       & 70.56          & 73.49          & 66.57          & 61.08          & 64.69          & 52.73          & 58.73          & 58.51          & -              \\
LAVT\cite{lavt} (CVPR'22)                 & Swin-B                                                                      & BERT                                                                     & 88M                                                                       & 72.73          & 75.82          & 68.79          & 62.14          & 68.38          & 55.10          & 61.24          & 62.09          & 60.50          \\
VLT\cite{vlt} (TPAMI'22)                 & Swin-B                                                                      & BERT                                                                     & 88M                                                                       & 72.96          & 75.96          & 69.60          & 63.53          & 68.43          & 56.92          & 63.49          & \underline{ 66.22}    & \underline{ 62.80}    \\
ReLA\cite{gres} (CVPR'23)                 & Swin-B                                                                      & BERT                                                                     & 88M                                                                       & 73.82          & 76.48          & 70.18          & \underline{ 66.04}    & \underline{ 71.02}    & \textbf{57.65} & \underline{ 65.00}    & 65.97          & 62.70          \\
DMMI\cite{dmmi} (ICCV'23)                 & Swin-B                                                                      & BERT                                                                     & 88M                                                                       & 74.13          & 77.13          & 70.16          & 63.98          & 69.73          & 57.03          & 63.46          & 64.19          & 61.98          \\
CGFormer\cite{cgformer} (CVPR'23)             & Swin-B                                                                      & BERT                                                                     & 88M                                                                       & 74.75          & 75.82          & 68.79          & 64.54          & 71.00          & 57.14          & 64.68          & 65.09          & 62.51          \\ 
MCRES\cite{mcres} (CVPR'23)                & Swin-B                                                                      & BERT                                                                     & 88M                                                                       & \underline{74.92}    & \underline{ 76.98}    & \underline{ 70.84}    & 64.32          & 69.68          & 56.64          & 63.51          & 64.90          & 61.63          \\ \hline \hline
\textbf{RISAM (Ours)}                   & VIT-H                                                                       & CLIP                                                                     & 6M                                                                        & \textbf{76.20} & \textbf{78.92} & \textbf{71.84} & \textbf{66.37} & \textbf{72.10} & \underline{ 57.33}    & \textbf{65.48} & \textbf{66.60} & \textbf{64.78} \\ 
\textbf{RISAM (Pr@90)  }                & VIT-H                                                                       & CLIP                                                                     & 6M                                                                        & 42.38          & 43.15          & 40.69          & 36.58          & 38.18          & 30.42          & 29.94          & 32.00          & 31.29          \\ \tabucline[1.2pt]{-}
\end{tabu}
\end{table*}

The next decoder block takes evolved language-aware visual feature $\overline{F}_{\mathrm{lav}}$ and evolved concatenated feature $\hat{F}_c$ from the previous layer as inputs. 

After that, the evolved language-aware visual feature $\overline{F}_{\mathrm{lav}}$ is up-sampled by two sequential blocks. Each consists of a transposed convolutional layer and a batch-normalization layer. We separate the evolved multi-modal query token $\hat{F}_m$ from the evolved concatenated feature $\hat{F}_c$, and send it to a MLP layer. Finally, we multiply the output of MLP with up-sampled visual features to generate the segmentation mask.

\subsection{Losses} 
In the training process, we adopt the linear combination of focal loss~\cite{lin2017focal} and dice loss~\cite{dice}. Denote $p$ as the possibility score of each pixel, and the focal loss is calculated as follows,
\begin{equation}
     \mathcal{L}_f = -(1-p)^\gamma\mathrm{log}(p),
\label{loss}
\end{equation}
where $\gamma$ is the hyper-parameters representing the importance of hard samples. Specifically, we set $\gamma$ as $2$.

Let $P$ denote the predicted mask, and $T$ is the target mask, the calculation of dice loss is shown as below,
\begin{equation}
     \mathcal{L}_d = 1-2\times \frac{\left | P\cap T \right | }{\left | P \right | +\left | T \right | },
\label{loss}
\end{equation}
where $\left | \cdot \right |$ symbolizes the sum of possibility scores.

Then, the whole loss function is formulated by
\begin{equation}
     \mathcal{L} = \omega_{f}\mathcal{L}_f + \omega_{d}\mathcal{L}_d,
\label{loss}
\end{equation}
where $\mathcal{L}_f$ and $\mathcal{L}_d$ are focal loss and dice loss, respectively. We set the weights ($\omega_f, \omega_d$) of two losses as $0.5$. 

\section{Experiments} 

\setlength{\tabcolsep}{5.5pt} 
\begin{table*}[t]
\centering
\caption{Performance comparison among each component of RISAM, including Feature Enhancement (FE), Mutual-Aware Attention (MA) blocks, and Mutual-Aware Mask Decoder (DE).}
\label{tab:big_abl}
\begin{tabu}{c|c|c|cccccc}
\tabucline[1.2pt]{-}
Settings      & Methods        & Datasets                       & Pr@50          & Pr@60          & Pr@70          & Pr@80          & Pr@90          & IoU            \\ \hline \hline
\# 1          & w/o FE         & \multirow{4}{*}{RefCOCO val}  & 77.83          & 75.09          & 71.92          & 64.51          & 37.84          & 69.19          \\
\# 2          & r. MA          &                               & 79.30          & 78.69          & 74.48          & 66.53          & 38.86          & 71.27          \\
\# 3          & r. DE          &                               & 84.69          & 82.45          & 78.82          & 69.69          & 40.04          & 74.97          \\
\textbf{\# 4} & \textbf{RISAM} &                               & \textbf{85.77} & \textbf{83.75} & \textbf{80.36} & \textbf{72.11} & \textbf{42.38} & \textbf{76.20} \\ \hline \hline
\# 1          & w/o FE         & \multirow{4}{*}{RefCOCO+ val} & 63.68          & 61.17          & 57.44          & 47.29          & 28.90          & 56.65          \\
\# 2          & r. MA          &                               & 69.53          & 66.44          & 62.15          & 53.03          & 29.71          & 61.83          \\
\# 3          & r. DE          &                               & 73.14          & 71.10          & 67.55          & 59.94          & 33.85          & 65.09          \\
\textbf{\# 4} & \textbf{RISAM} &                               & \textbf{74.59} & \textbf{72.69} & \textbf{69.32} & \textbf{61.87} & \textbf{36.58} & \textbf{66.37} \\ \hline \hline
\# 1          & w/o FE         & \multirow{4}{*}{G-Ref val (U)} & 65.15          & 62.42          & 57.78          & 48.97          & 26.70          & 58.58          \\
\# 2          & r. MA          &                               & 67.52          & 64.16          & 60.18          & 51.06          & 27.40          & 60.21          \\
\# 3          & r. DE          &                               & 71.32          & 68.28          & 63.16          & 54.68          & 28.26          & 63.84          \\
\textbf{\# 4} & \textbf{RISAM} &                               & \textbf{73.88} & \textbf{70.30} & \textbf{65.89} & \textbf{56.47} & \textbf{29.94} & \textbf{65.48} \\ \tabucline[1.2pt]{-}
\end{tabu}
\begin{tablenotes}
     \item[1] (r.: replace with other modules.)
\end{tablenotes}
\end{table*}

\subsection{Datasets}

We conduct experiments on three widely used RIS datasets, including RefCOCO \& RefCOCO+~\cite{refcoco}, and G-Ref~\cite{GRef}. Besides, we also conduct experiments on PhraseCut~\cite{wu2020phrasecut} and gRefCOCO~\cite{gres} to evaluate the generalization ability and multi-object RIS performance, respectively. The details of these five datasets are described as follows.

\subsubsection{Three Common RIS Benchmarks}
RefCOCO and RefCOCO+~\cite{refcoco} are two of the largest and most well-known datasets for referring image segmentation, which were collected using a two-player game. To be specific, there are 19,994 images with 142,209 referring expressions describing 50,000 objects in the RefCOCO dataset and 19,992 images with 141,564 expressions describing 49,856 objects in the RefCOCO+ dataset. Expressions of RefCOCO have an average length of 3.61, while that of RefCOCO+ have an average length of 3.53. Compared with RefCOCO, expressions about absolute locations, $i.e.$, left/right, are forbidden in RefCOCO+.

G-Ref~\cite{GRef} contains 104,560 referring expressions for 54,822 objects in 26,711 images. Unlike the above two datasets, the language expression of G-Ref is collected by using a non-interactive setting. The average length of G-Ref expressions is 8.4 words. It is worth mentioning that there are two versions of the G-Ref dataset in terms of the split methods. One is the UMD split, which consists of both validation and testing sets. The other is the Google split, which only contains a testing set.

\subsubsection{Generalization Ability Benchmark} PhraseCut~\cite{wu2020phrasecut} is composed of 77,262 images and 345,486 phrase-region pairs with 1,272 categories. These images occupy around $70\%$ Visual Genome. For the evaluation of the generalization ability, we use the test split, which contains 2,545 images and 14,354 expressions.

\subsubsection{Multi-object Segmentation Benchmark} gRefCOCO~\cite{gres} is a novel referring image segmentation dataset for multi-object segmentation. It contains 278,232 expressions, which refer to 60,287 distinct instances in 19,994 images. Some expressions are inherited from RefCOCO.

\subsection{Implementation Details}
We utilize VIT pre-trained by SAM~\cite{sam} and GPT-2.0 pre-trained by CLIP~\cite{clip} as the image encoder and text encoder, respectively. Notably, these encoders are frozen during the training process. For the input images, we follow the image pre-processing of SAM~\cite{sam}. For RefCOCO, we set the maximum word number to 17. For RefCOCO+ and G-Ref, we set it to 25. In the training process, we use Adam~\cite{loshchilov2017decoupled} optimizer with the initial learning rate of $1\textrm{e}^{-4}$ and train the network for 50 epochs. The learning rate is decreased by the factor 0.1 at the 30th and 42nd epochs. This model is trained with a batch size of 32 on 8 Nvidia A100 SXM4 with 40G memory.

In the inference process, we up-sample the mask to the original size. Then, we binarize the mask with a threshold of $0.35$. No other post-processing tools are used.

\subsection{Metrics}
Following previous works~\cite{vlt,gres}, we utilize two metrics in our experiments, including mask Intersection-over-Union (IoU) score and Precision with thresholds (Pr@$X$). Specifically, IoU scores reveal the predicted mask quality by calculating intersection regions over union regions between the predicted mask and the ground truth across all testing samples. Besides, Pr@$X$ denotes the ratio of predicted masks with IoU scores higher than the threshold $X\in \{50,60,70,80,90\}$. For example, Pr@$50$ denotes the location ability of the model, while Pr@$90$ shows the ability to generate a high-quality mask.

\subsection{Comparison With State-of-the-art Methods}
We compare the proposed RISAM with previous state-of-the-art (SOTA) methods on the three most widely used benchmarks, $i.e.$, RefCOCO, RefCOCO+, and G-Ref. Quantitative results are shown in Tab.~\ref{tab:comp}.

Our method achieves significant improvements over the second-best SOTA method, MCRES~\cite{mcres}, on the RefCOCO dataset. Specifically, our method outperforms MCRES by $1.28\%$, $1.94\%$, and $1.00\%$ on the val, testA, and testB split, respectively. These results demonstrate the effectiveness of our framework for the RIS task. 

On the RefCOCO+ dataset, our RISAM improves over ReLA~\cite{gres} on the val and testA splits by $0.33\%$ and $1.08\%$, respectively. However, we observe a slight performance drop of $0.32\%$ on the testB split compared to ReLA. A possible reason is that the frozen text encoder gets sub-optimal linguistic feature representation for language expression without absolute locations. When the test set ($i.e.$, testB split) contains images with multiple objects that are hard to be distinguished without absolute locations, our method exhibits inferior performance.


Finally, on another more complex G-Ref dataset, our method achieves an IoU improvement of $0.48\%$, $0.38\%$, and $1.98\%$ on the val (U), test (U), and val (G) split, respectively. This improvement indicates that our method is also competitive for long and causal language expressions. Besides, we also demonstrate the ratio of predicted masks with IoU scores higher than $90\%$. According to the last row of Tab.~\ref{tab:comp}, our method typically segments a high-quality mask.

\subsection{Ablation Study}
 To verify the effectiveness of the proposed modules of our method, we conduct ablation studies to investigate each component, including Feature Enhancement (FE), Mutual-Aware Attention (MA), and Mutual-Aware Mask Decoder (DE) on the RefCOCO val dataset, as shown in Tab.~\ref{tab:big_abl}. For setting \# 2, we replace our Mutual-Aware Attention Block with convolutional layers, which receive the concatenation of visual and linguistic features as inputs. For setting \# 3, we use the SAM's decoder~\cite{sam} for the variant excluding the proposed decoder.

\subsubsection{Mutual-Aware Attention Blocks}
Mutual-Aware Attention blocks are introduced to weight different image regions and different words in the sentence. It brings an IoU improvement by $4.93\%/4.54\%/5.27\%$ on three different benchmarks. We conduct three sets of experiments to get a comprehensive understanding of our Mutual-Aware Attention.

\begin{figure}[t]
	\centering{\includegraphics[scale=0.24]{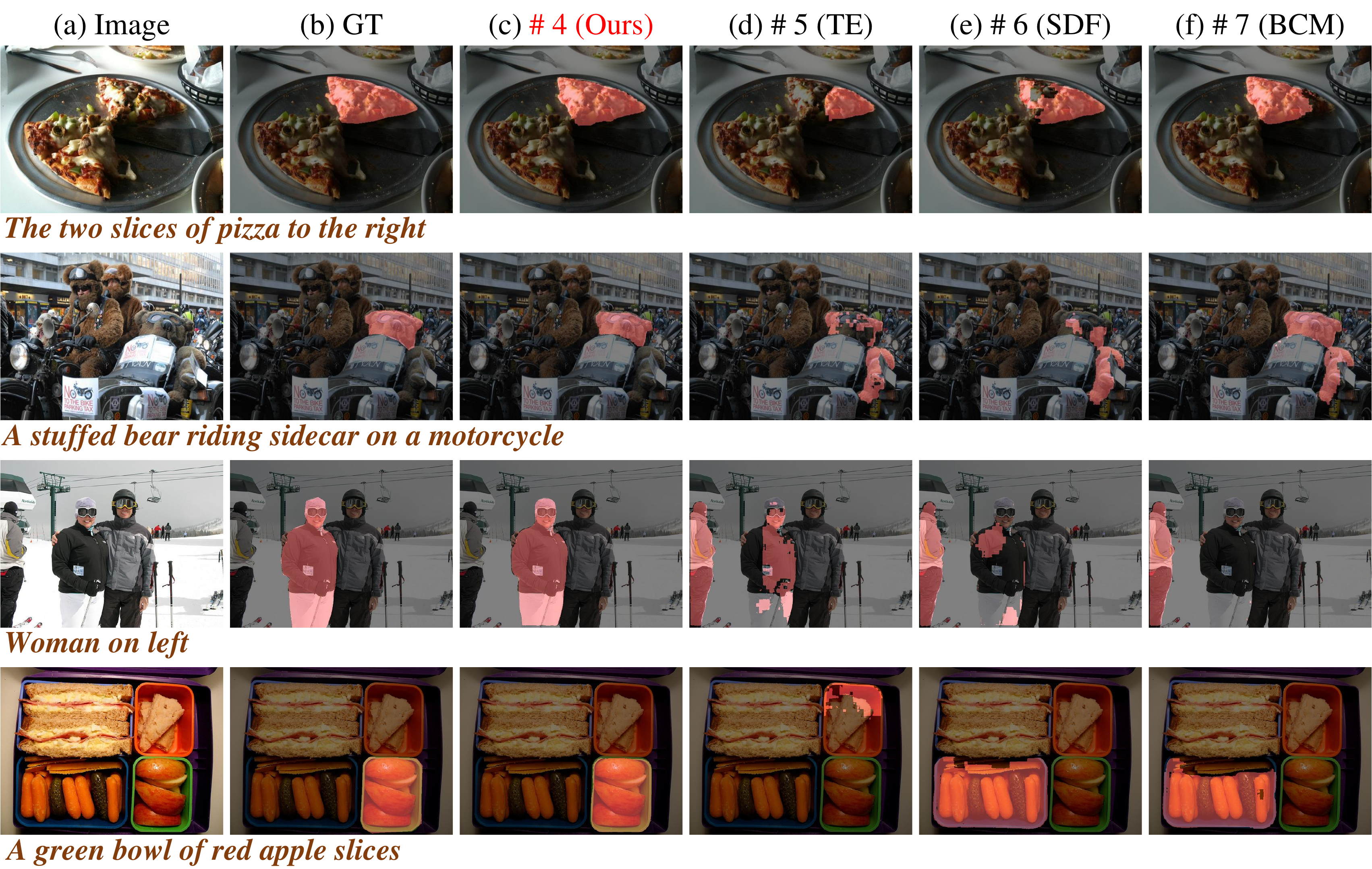}}
	\caption{ Visualized examples of the ablation study of Mutual-Aware Attention. (a) The input image. (b) Ground Truth (GT). (c) Our method. (d) Transformer encoder (TE) in ResTR. (e) Spatial Dynamic Fusion (SDF) in VLT. (f) Bidirectional cross-modal module (BCM) in BRINet.}
    \label{pic:abl_vla}
\end{figure}

\textbf{Comparison with other fusion methods.} To verify the superiority of Mutual-Aware Attention, we conduct experiments that use other methods~\cite{kim2022restr,vlt, BRINET} to incorporate features of different modalities. Specifically, ReSTR~\cite{kim2022restr} utilizes a transformer encoder (TE) to model the long-range dependencies. VLT~\cite{vlt} adopts the Spatial Dynamic Fusion (SDF) to produce different linguistic feature vectors, which is equivalent to using only Visual-Guided Attention. BRINet~\cite{BRINET} introduces a serial bidirectional cross-modal (BCM) module to utilize visual and linguistic guidance.

\setlength{\tabcolsep}{5pt} 
\begin{table*}[t]
\centering
\caption{The comparison between our MA and other methods, including TE in ReSTR, SDF in VLT, BCM in BRINet, and without attention mask.}
\label{tab:abl_vla}
\begin{tabu}{c|c|c|cccccc}
\tabucline[1.2pt]{-}
Settings      & Methods            & Datasets                   & Pr@50          & Pr@60          & Pr@70          & Pr@80          & Pr@90          & IoU            \\ \hline \hline
\# 5          & TE (ReSTR)         & \multirow{5}{*}{RefCOCO val}  & 83.94          & 81.29          & 78.62          & 70.17          & 39.77          & 74.85          \\
\# 6          & SDF (VLT)          &                           & 84.45          & 82.38          & 78.91          & 70.25          & 40.22          & 75.08          \\
\# 7          & BCM (BRINet)       &                           & 84.63          & 82.60          & 79.17          & 70.54          & 40.56          & 75.20          \\
\# 8          & w/o attn.mask      &                           & 85.01          & 83.14          & 80.08          & 71.62          & 41.71          & 75.52          \\
\textbf{\# 4} & \textbf{MA (Ours)} &                           & \textbf{85.77} & \textbf{83.75} & \textbf{80.36} & \textbf{72.11} & \textbf{42.38} & \textbf{76.20} \\ \hline \hline
\# 5          & TE (ReSTR)         & \multirow{5}{*}{RefCOCO+ val} & 71.51          & 69.47          & 66.02          & 55.41          & 28.32          & 63.23          \\
\# 6          & SDF (VLT)          &                           & 72.14          & 70.10          & 66.55          & 56.94          & 28.95          & 64.09          \\
\# 7          & BCM (BRINet)       &                           & 73.51          & 70.89          & 66.48          & 57.41          & 30.82          & 64.73          \\
\# 8          & w/o attn.mask      &                           & 73.81          & 71.74          & 67.36          & 58.31          & 32.61          & 65.82          \\
\textbf{\# 4} & \textbf{MA (Ours)} &                           & \textbf{74.59} & \textbf{72.69} & \textbf{69.32} & \textbf{61.87} & \textbf{36.58} & \textbf{66.37} \\ \hline \hline
\# 5          & TE (ReSTR)         & \multirow{5}{*}{G-Ref val (U)}    & 68.96          & 64.94          & 59.64          & 50.58          & 24.78          & 62.21          \\
\# 6          & SDF (VLT)          &                           & 70.56          & 66.44          & 61.48          & 51.94          & 26.80          & 63.17          \\
\# 7          & BCM (BRINet)       &                           & 70.92          & 67.46          & 63.20          & 54.98          & 28.48          & 63.73          \\
\# 8          & w/o attn.mask      &                           & 73.62          & 69.98          & 65.20          & 55.74          & 29.17          & 64.83          \\
\textbf{\# 4} & \textbf{MA (Ours)} &                           & \textbf{73.88} & \textbf{70.30} & \textbf{65.89} & \textbf{56.47} & \textbf{29.94} & \textbf{65.48} \\ \tabucline[1.2pt]{-}
\end{tabu}
\end{table*}

\begin{table}[ht]
\setlength{\tabcolsep}{6pt} 
\centering
\caption{Ablation study of the number of Mutual-Aware Attention blocks.}
\label{tab:VLA}
\begin{tabu}{c|cccccc}
\tabucline[1.2pt]{-}
\multicolumn{1}{l|}{MA Num} & Pr@50    & Pr@60    & Pr@70                     & Pr@80                     & Pr@90                     & IoU                       \\ \hline \hline
1   & 84.45                     & 82.94               & 79.95                     & 71.69                     & 42.51                     & 75.88                     \\
2    & \textbf{85.77}      & \textbf{83.75}              & \textbf{80.36}                     & \textbf{72.11}                    & \textbf{42.38}                 & \textbf{76.20}                     \\
3        & 84.32                     & 82.76                    & 79.51                     & 71.46                     & 41.16                     & 75.89                     \\ \tabucline[1.2pt]{-}
\end{tabu}
\end{table}

\begin{table*}[t]
\setlength{\tabcolsep}{5.5pt} 
\centering
\caption{Ablation study of the Mutual-Aware Mask Decoder.}
\label{tab:mask_decoder}
\begin{tabu}{c|c|c|cccccc}
\tabucline[1.2pt]{-}
Settings      & Methods       & Datasets                       & Pr@50          & Pr@60          & Pr@70          & Pr@80          & Pr@90          & IoU            \\ \hline \hline
\# 9          & w/ TOI-A      & \multirow{3}{*}{RefCOCO val}  & 84.76          & 82.63          & 79.21          & 71.07          & 40.65          & 75.40          \\
\# 10         & w/o ELG       &                               & 83.79          & 81.19          & 78.55          & 67.96          & 38.62          & 74.42          \\
\textbf{\# 4} & \textbf{Ours} &                               & \textbf{85.77} & \textbf{83.75} & \textbf{80.36} & \textbf{72.11} & \textbf{42.38} & \textbf{76.20} \\ \hline \hline
\# 9          & w/ TOI-A      & \multirow{3}{*}{RefCOCO+ val} & 72.96          & 70.88          & 67.70          & 59.32          & 33.41          & 64.87          \\
\# 10         & w/o ELG       &                               & 71.46          & 69.30          & 66.59          & 56.64          & 31.75          & 63.45          \\
\textbf{\# 4} & \textbf{Ours} &                               & \textbf{74.59} & \textbf{72.69} & \textbf{69.32} & \textbf{61.87} & \textbf{36.58} & \textbf{66.37} \\ \hline \hline
\# 9          & w/ TOI-A      & \multirow{3}{*}{G-Ref val (U)} & 71.14          & 67.26          & 62.38          & 54.28          & 28.36          & 63.84          \\
\# 10         & w/o ELG       &                               & 70.30          & 67.02          & 62.68          & 53.98          & 28.24          & 62.98          \\
\textbf{\# 4} & \textbf{Ours} &                               & \textbf{73.88} & \textbf{70.30} & \textbf{65.89} & \textbf{56.47} & \textbf{29.94} & \textbf{65.48} \\ \tabucline[1.2pt]{-}
\end{tabu}
\end{table*}

According to Tab.~\ref{tab:abl_vla}, our MA outperforms TE, SDF, BCM by $1.35\%/3.14\%/3.27\%$, $1.12\%/2.28\%/2.31\%$, $1.00\%/1.64\%/1.75\%$ IoU scores, respectively. This is because existing methods only explore the informative words of each image region, while our method also provides the corresponding image regions of each word in the language expression and generates vision-aware linguistic features. We also observe that our method exceeds a significantly greater performance improvement on more challenging datasets compared to other methods. For example, on RefCOCO+, where the location words are absent in these expressions, our method gets more improvement than RefCOCO, which indicates our method tends to get a more comprehensive vision-language understanding.
Finally, according to \# 8, the attention mask alleviates the interference of irrelevant pairs between visual and linguistic features, which further improves the performance by $0.68\%/0.55\%/0.65\%$ IoU. We also list some
visualized examples of different methods in Fig.~\ref{pic:abl_vla}.

\begin{figure}[t]
	\centering{\includegraphics[scale=0.3]{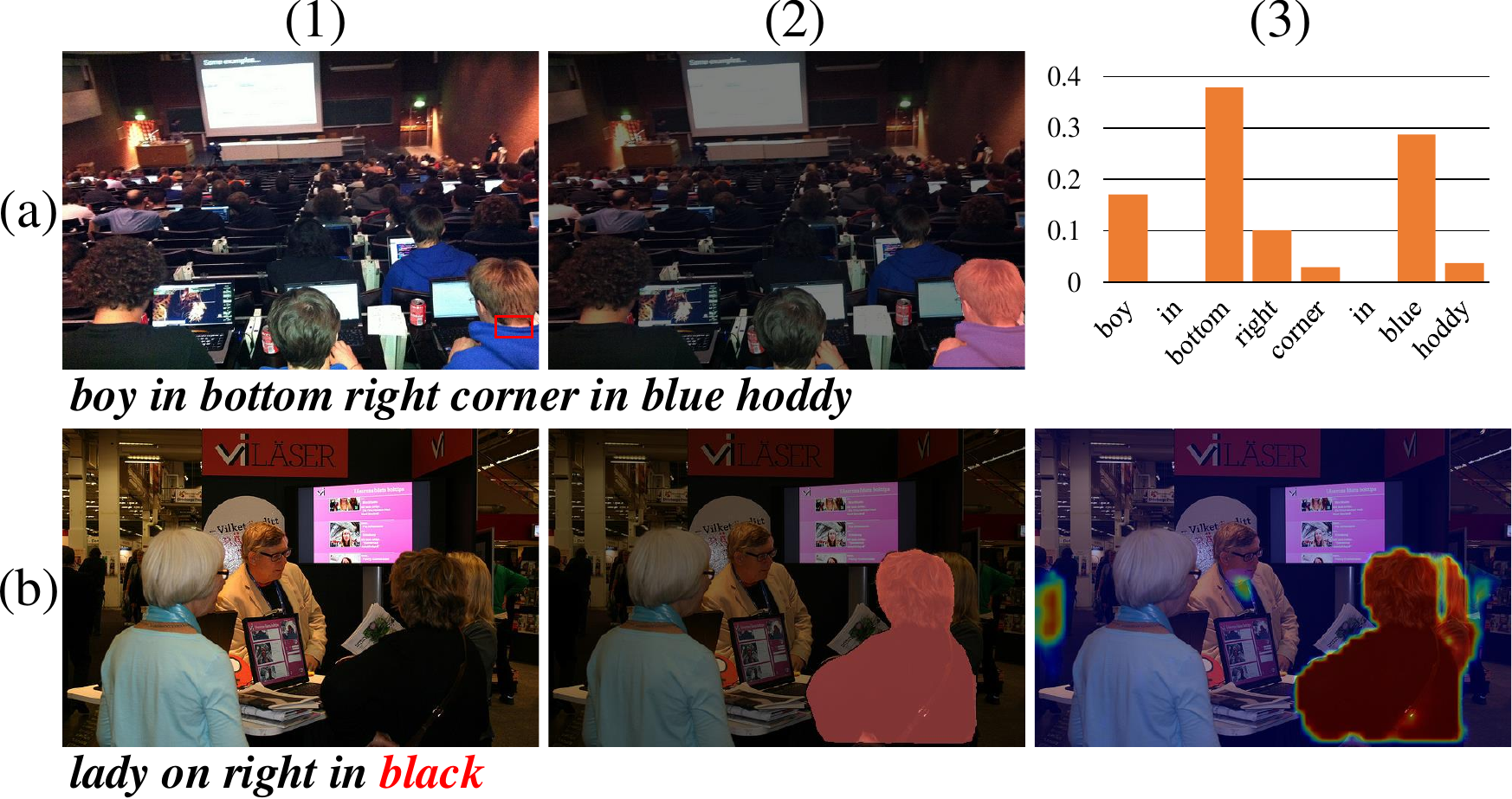}}
	\caption{Results of Vision-Guided Attention (a) and Language-Guided Attention (b).}
    \label{pic:vla_result}
\end{figure}

\textbf{The outputs of Mutual-Aware Attention blocks.} 
We visualize the outputs of Vision-Guided Attention and Language-Guided Attention in Fig.~\ref{pic:vla_result}(a) and (b), respectively. For the red rectangle in Fig.~\ref{pic:vla_result}(a1), we list attention weights of each word in Fig.~\ref{pic:vla_result}(a3). Our model considers \emph{`bottom'} and \emph{`blue'} as the most informative words. Thus, our prediction mask accurately locates the bottom boy in blue, as shown in Fig.~\ref{pic:vla_result}(a2). Similarly, for the word \emph{`black'}, we show its attention map in Fig.~\ref{pic:vla_result}(b3). In the mask decoder, the final segmentation mask is refined according to the attention map. Our prediction mask is shown in Fig.~\ref{pic:vla_result}(b2).

\begin{figure}[t]
	\centering{\includegraphics[scale=0.66]{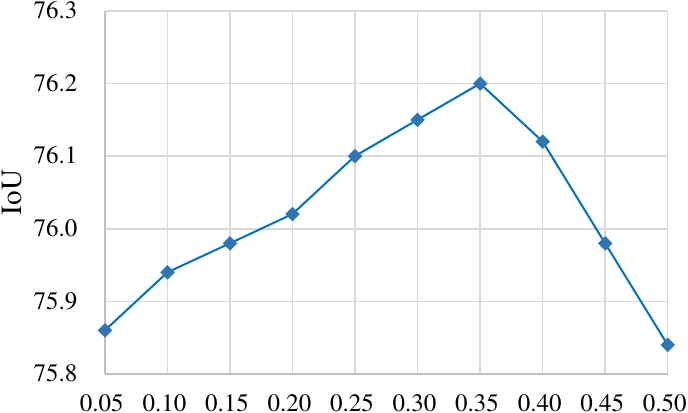}}
	\caption{ IoU performance about the value of $\tau$.} 
 \label{pic:zhexian}
\end{figure}

\textbf{The value of $\tau$.}
We conduct experiments to select the value of $\tau$. To be specific, we set the parameter $\tau$ range between 0.05 and 0.5, with increments of 0.05. As shown in Fig.~\ref{pic:zhexian}, when the value is too small, irrelevant pairs between visual and linguistic features lead to performance degradation. Conversely, when the value of $\tau$ is too big, some necessary visual/linguistic features are masked. As a result, the IoU performance decreases. When the value is $0.35$, our method gains the best performance. Therefore, we set the value of $\tau$ as $0.35$.

\textbf{The number of Mutual-Aware Attention blocks.}
The number of Mutual-Aware Attention Blocks may influence the performance of our method. We conduct experiments on the selection of the number of Mutual-Aware Attention Blocks, as shown in Tab.~\ref{tab:VLA}. When the number is $2$, our method achieves the best performance. Therefore, we build two Mutual-Aware Attention blocks for our method.

\subsubsection{Mutual-Aware Mask Decoder}
According to Tab.~\ref{tab:big_abl}, replacing our mask decoder with SAM's decoder reduces the IoU score by $1.23\%/1.28\%/1.64\%$. This reduction is caused by the \emph{token to image attn.}. The detailed analysis is as follows.

\textbf{The effectiveness of TOI-A layer.} (TOI-A) layer in SAM's decoder. Specifically, the TOI-A layer performs cross-attention by taking prompt tokens containing output tokens and linguistic features as queries (Q), visual features as keys (K), and vectors (V). Since these output tokens are initialized randomly, they make an uncertain adjustment to the evolved visual features and thus affect the performance. To verify the disadvantage of TOI-A layer for RIS, we insert this layer into each block of our decoder. As shown in \# 9 of Tab.~\ref{tab:mask_decoder}, the TOI-A layer leads to $0.80\%/1.50\%/1.64\%$ IoU decrease. 

\textbf{The effectiveness of extra linguistic guidance.} To verify the effectiveness of extra linguistic guidance (ELG), we also implement the experiment without extra linguistic guidance (\# 10). Specifically, we only send language-aware visual features into the mask decoder, where the structure is the same as~\cite{vlt}. \# 10 in Tab.~\ref{tab:mask_decoder} indicates that extra linguistic guidance improves the IoU performance by $1.78\%/2.92\%/2.50\%$, which demonstrates that our extra linguistic guidance guides the model to generate segmentation masks more corresponding to the referring expressions.

\subsubsection{Feature Enhancement}
As shown in Tab.~\ref{tab:abl_fm}, the Feature Enhancement module significantly improves the performance of RISAM by $7.01\%$ IoU score. To understand Feature Enhancement concretely, we conduct the experiment by using another well-known backbone-adaption baseline. Specifically, we adopt VIT-DET~\cite{vitdet} as the compared baseline, which uses only the feature map from the last layer of the backbone to generate multi-scale features. The quantitative evaluations are shown in Tab.~\ref{tab:abl_fm}.

\begin{table}[ht]
\setlength{\tabcolsep}{4.5pt} 
\centering
\caption{ Ablation study of the Feature Enhancement.}
\label{tab:abl_fm}
\begin{tabu}{c|ccccccc}
\tabucline[1.2pt]{-}
Settings & Methods & \multicolumn{1}{l}{Pr@50} & \multicolumn{1}{l}{Pr@60} & \multicolumn{1}{l}{Pr@70} & \multicolumn{1}{l}{Pr@80} & \multicolumn{1}{l}{Pr@90} & \multicolumn{1}{c}{IoU} \\ \hline \hline
\# 1    & w/o FE  & 77.83          & 75.09        & 71.92                     & 64.51                     & 37.84                     & 69.19 \\
\# 11    & VIT-DET & 80.72         & 77.40       & 74.74         & 66.79        & 37.85                     & 71.55                   \\ \hline \hline
\textbf{\# 4}    & \textbf{Ours}  & \textbf{85.77}           & \textbf{83.75}  & \textbf{80.36}           & \textbf{72.11}           & \textbf{42.38}           & \textbf{76.20}         \\ \tabucline[1.2pt]{-}
\end{tabu}
\end{table}

Compared with removing Feature Enhancement (FE) module (\# 1), multi-scale features generated from the last layer improve the performance by $2.36\%$. However, compared with using features from different layers, this baseline shrinks the IoU performance by $4.65\%$. The reason for performance degradation is that features from the last layer contain highly global information, and multi-scale features generated from the last layer exhibit a limited representation of grained details that are essential for RIS.

\subsection{Replacement of Different Visual Backbones}
There is a concern that the improvement of our method comes from the utilization of SAM. To alleviate this concern, we train our models with different visual backbones, $i.e.$, VIT-B and Swin-B. The IoU performance is recorded in Tab.~\ref{tab:visual}. For ViT-B, our RISAM exceeds ETRIS by approximately $1.5\%$ IoU on RefCOCO. For Swin-B, our RISAM also improves around $1\%$ IoU over ReLA. These improvements come from our specific designs. First, to use extra linguistic guidance in the mask decoder, we first design the Mutual-Aware Attention mechanism to get multi-modal features, which contain language-aware visual features and vision-aware linguistic features. Then, the vision-aware linguistic features act as the extra guidance for language-aware visual features in the mask decoder. Second, we introduce the parameter-efficient fine-tuning framework into RIS, which inherits the power of SAM.

\begin{table}[ht]
    \centering
    \setlength{\tabcolsep}{10pt} 
    \caption{IoU comparison across different visual backbones.}
    \label{tab:visual}
    \begin{tabu}{c|c|ccc}
        \tabucline[1.2pt]{-}
        \multirow{2}{*}{Methods} & \multirow{2}{*}{Backbones} & \multicolumn{3}{c}{RefCOCO}                      \\ \cline{3-5} 
                                 &                            & val            & testA          & testB          \\ \hline \hline
    ReSTR (CVPR'22)                    & ViT-B                      & 67.22          & 69.30          & 64.45          \\
    ETRIS (ICCV’23)                    & ViT-B                      & 70.51          & 73.51          & 66.63          \\
        \textbf{RISAM (Ours)}                    & ViT-B                     & \textbf{72.16} & \textbf{74.73} & \textbf{68.25} \\ \hline \hline
    VLT (TPAMI'22)                     & Swin-B                     & 72.96          & 75.96          & 69.60          \\
    ReLA (CVPR'23)                    & Swin-B                      & 73.82          & 76.48          & 70.18          \\
        \textbf{RISAM (Ours)}                   & Swin-B                     & \textbf{74.96} & \textbf{77.51} & \textbf{70.96} \\ \hline \hline
        \textbf{RISAM (Ours)}                   & SAM                        & \textbf{76.20} & \textbf{78.92} & \textbf{71.84} \\ \tabucline[1.2pt]{-}
        \end{tabu}
\end{table}

\subsection{Generalization Ability}
To demonstrate the generalization ability of our method, we conduct experiments on the test split of PhraseCut~\cite{wu2020phrasecut}. PhraseCut contains 1287 categories, which is much more diverse than 80 categories in COCO. Thus, we compare with two previous methods on PhraseCut to evaluate their generalization ability.   
\begin{table}[ht]
\setlength{\tabcolsep}{19pt} 
\centering
\caption{Generalization ability of different methods.}
\label{tab:zero-shot}
\begin{tabu}{c|ccc}
\tabucline[1.2pt]{-}
\multirow{2}{*}{Training Set} & \multicolumn{3}{c}{IoU results on PhraseCut}        \\ \cline{2-4} 
                              & CRIS  & LAVT  & \textbf{Ours}   \\ \hline \hline
RefCOCO                       & 15.53 & 16.68 & \textbf{22.82} \\
RefCOCO+                      & 16.30 & 16.64 & \textbf{21.68} \\
G-Ref                         & 16.24 & 16.05 & \textbf{22.47} \\ \tabucline[1.2pt]{-}
\end{tabu}
\end{table}

\begin{figure}[ht]
	\centering{\includegraphics[scale=0.24]{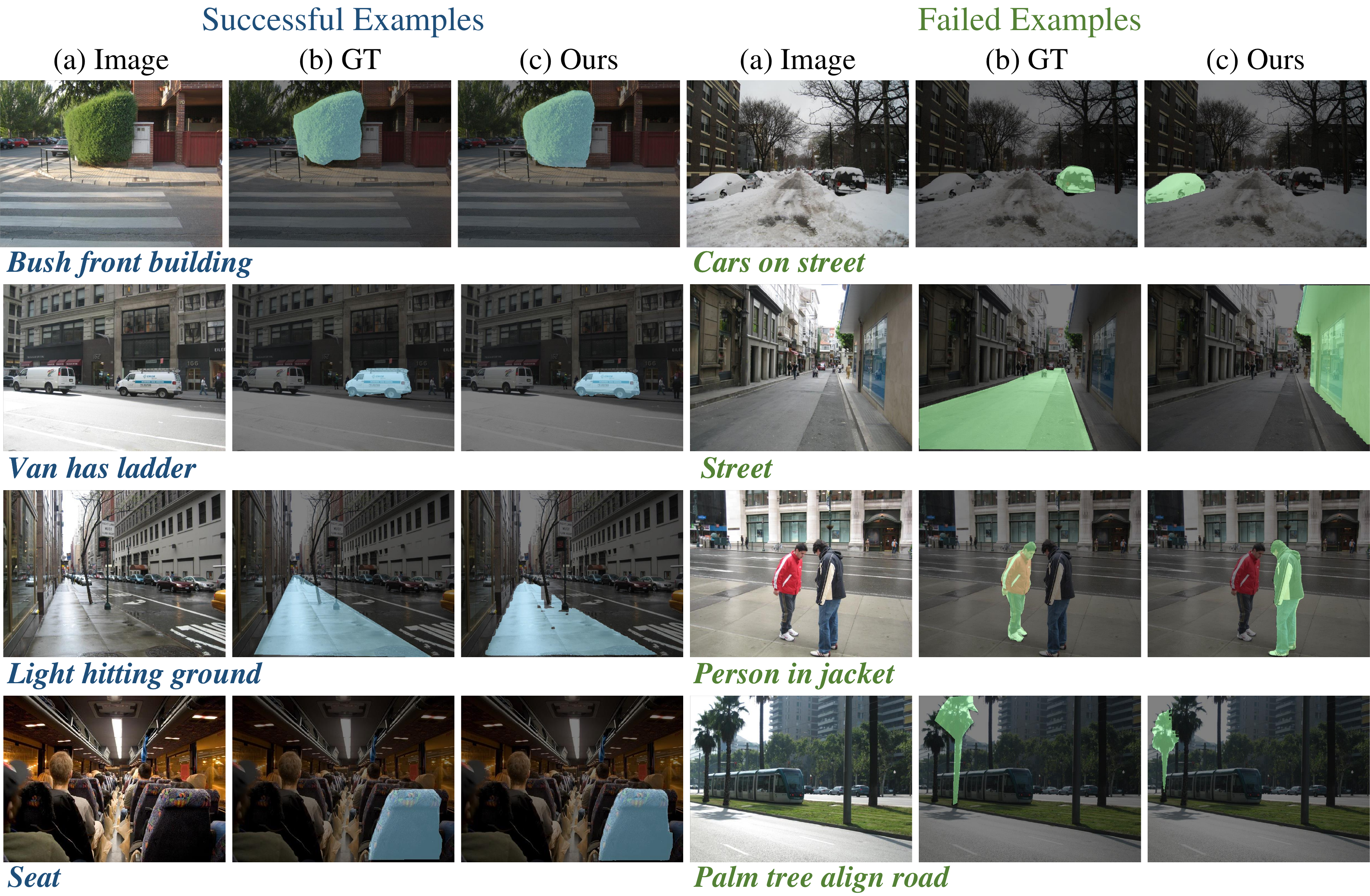}}
	\caption{ The successful examples (left) and failed examples (right) in PhraseCut.}
    \label{pic:phrase}

\end{figure}

As shown in Tab.~\ref{tab:zero-shot}, our method surpasses previous methods in terms of generalization ability. For example, when training on the RefCOCO dataset, our method exceeds CRIS and LAVT by $7.29\%$ and $6.14\%$, respectively. This advantage comes from the PEFT framework and our specific designs (mentioned in the above subsection). In contrast, encoders of other methods are trainable and thus might be biased to fine-tuned datasets. We also provide some successful and failed visualized examples in Fig.~\ref{pic:phrase}. For expressions with clear references, $e.g.$, \emph{`Van has ladder'}, our method is capable of accurately segmenting objects. However, our method may segment the wrong object when the expression is ambiguous, such as \emph{`Cars on street'}.

\subsection{Unfreezing the Text Encoder}
As shown in \# 3 in Tab.~\ref{tab:freeze}, the unfrozen text encoder leads to IoU drops of different levels on three benchmarks. Besides, the IoU performance on the PharseCut significantly decreases from $22.82\%$ to $17.96\%$. This is because the number of RIS datasets is insufficient for the convergence of  CLIP, where the text encoder is trained on 400 million image-text pairs.

\begin{table}[ht]
\centering
\setlength{\tabcolsep}{10pt} 
\caption{IoU comparison across different visual backbones.}
\label{tab:freeze}
\begin{tabu}{c|c|ccc}
\tabucline[1.2pt]{-}
Datasets                   & Splits   & n. IoU & IoU   & Decline                     \\ \hline \hline
                           & val      & 76.20  & 75.82 & \textcolor{blue}{0.38 $\downarrow$} \\
                           & testA    & 78.92  & 77.24 & \textcolor{blue}{1.68 $\downarrow$} \\
\multirow{-3}{*}{RefCOCO}  & testB    & 71.84  & 70.06 & \textcolor{blue}{1.78 $\downarrow$} \\ \hline \hline
                           & val      & 66.37  & 64.32 & \textcolor{blue}{2.05 $\downarrow$} \\
                           & testA    & 72.10  & 71.38 & \textcolor{blue}{0.72 $\downarrow$} \\
\multirow{-3}{*}{RefCOCO+} & testB    & 57.33  & 56.42 & \textcolor{blue}{0.91 $\downarrow$} \\ \hline \hline
                           & val (U)  & 65.48  & 64.19 & \textcolor{blue}{1.29 $\downarrow$} \\
                           & test (U) & 66.60  & 64.80 & \textcolor{blue}{1.80 $\downarrow$} \\
\multirow{-3}{*}{G-Ref}    & val (G)  & 64.78  & 62.45 & \textcolor{blue}{2.33 $\downarrow$} \\ \hline \hline
PhraseCut                  & test     & 22.82  & 17.96 & \textcolor{blue}{4.86 $\downarrow$} \\ \tabucline[1.2pt]{-}
\end{tabu}
\end{table}

\subsection{Performance on Multi-object Referring Image Segmentation}

\setlength{\tabcolsep}{12pt} 
\begin{table}[h]
\centering
\caption{Performance comparison among different methods on multi-object referring image segmentation in terms of IoU (\%).}
\label{tab:multi-object}
\begin{tabu}{c|cccc}
\tabucline[1.2pt]{-}
Methods      & val             & testA           & testB           & Avg.            \\ \hline \hline
MattNet~\cite{yu2018mattnet}      & 47.51           & 58.66           & 45.33           & 50.50           \\
LTS~\cite{lts}          & 52.30           & 61.87           & 49.96           & 54.71           \\
VLT~\cite{vlt}          & 52.51           & 62.19           & 50.52           & 55.07           \\
CRIS~\cite{cris}         & 55.34           & 63.82           & 51.04           & 56.73           \\
LAVT~\cite{lavt}         & 57.64           & 65.32           & 55.04           & 59.33           \\
ReLA~\cite{gres}         & 62.42           & 69.26           & 59.88           & 63.85           \\ \hline
\textbf{RISAM (Ours)} & \textbf{66.52} & \textbf{70.01} & \textbf{62.23} & \textbf{66.25} \\ \tabucline[1.2pt]{-}
\end{tabu}
\end{table}

\begin{figure}[h]
	\centering{\includegraphics[scale=0.24]{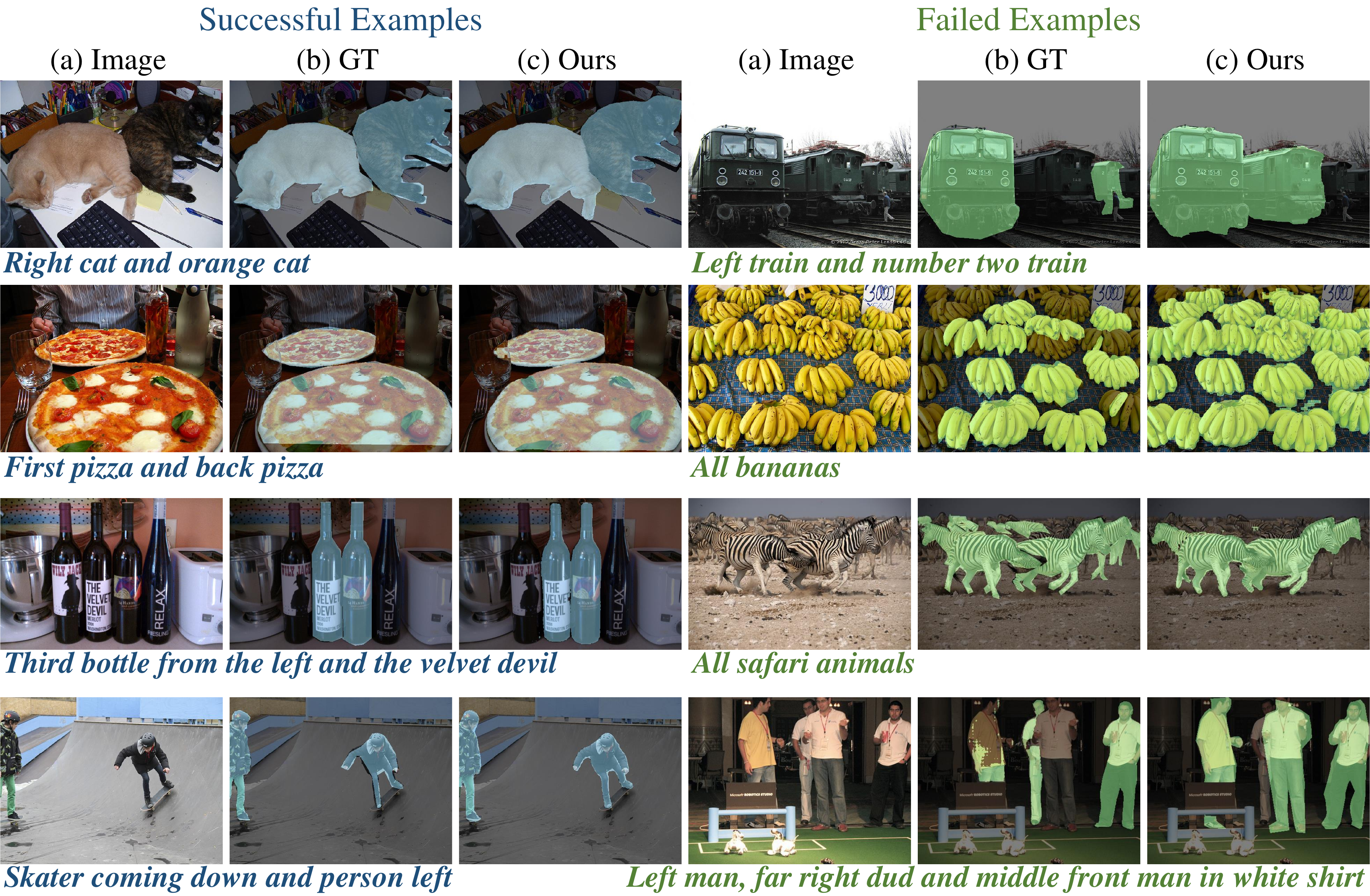}}
	\caption{ The successful examples (left) and failed examples (right) of multi-object referring image segmentation.}
    \label{pic:gres}
\end{figure}

Recently, Liu et al.~\cite{gres} build a new RIS dataset called gRefCOCO. Different from conventional RIS datasets, this dataset contains both multi-target and single-target expressions. To assess the performance of our method in multi-target RIS, we conduct experiments on this novel benchmark. As illustrated in Tab.~\ref{tab:multi-object}, our model surpasses the second SOTA method, $i.e.$, ReLA, by $3.10\%/0.75\%/2.35\%$ on val/testA/testB splits. This significant improvement indicates that our method is also competitive on multi-target RIS. In Fig.~\ref{pic:gres}, we demonstrate some successful examples and failed examples of gRefCOCO. For the expression \emph{`all'}, our method may miss some objects (shown in \emph{`all bananas'} and \emph{`all safari animals'}). For the expression \emph{`number two'}, our method find the \emph{`number two'} from the right instead of the \emph{`number two'} from the left. In the future, we will improve the performance of our model to overcome these faults.

\section{Discussions}
Through a large amount of experiments, we conclude some insights about adapting SAM into other tasks, which may inspire other research.

\setlength{\tabcolsep}{10pt} 
\begin{table}[ht]
\centering
\caption{IoU performance of SAM and our RISAM.}
\label{tab:sam}
\begin{tabu}{c|c|ccc}
\tabucline[1.2pt]{-}
Datasets                  & Splits   & SAM   & RISAM & Increase \\ \hline \hline
\multirow{3}{*}{RefCOCO}  & val      & 57.23 & 76.20 & \textcolor{red}{18.97 $\uparrow$}    \\
                          & testA    & 58.71 & 78.92 & \textcolor{red}{20.21 $\uparrow$}    \\
                          & testB    & 56.34 & 71.84 & \textcolor{red}{15.50 $\uparrow$}    \\ \hline \hline
\multirow{3}{*}{RefCOCO+} & val      & 44.95 & 66.37 & \textcolor{red}{21.42 $\uparrow$}    \\
                          & testA    & 50.35 & 72.10 & \textcolor{red}{21.75 $\uparrow$}    \\
                          & testB    & 41.38 & 57.33 & \textcolor{red}{15.95 $\uparrow$}    \\ \hline \hline
\multirow{3}{*}{G-Ref}    & val (U)  & 47.87 & 65.48 & \textcolor{red}{17.61 $\uparrow$}    \\
                          & test (U) & 49.31 & 66.60 & \textcolor{red}{17.29 $\uparrow$}    \\
                          & val (G)  & 46.23 & 64.78 & \textcolor{red}{18.55 $\uparrow$}    \\
                          \tabucline[1.2pt]{-}
\end{tabu}
\end{table}

We conduct experiments about directly adapting SAM into RIS. Specifically, we freeze the visual and text encoder and only train the linear layer (ensure consistency of feature dimensions across different modalities) and the mask decoder. The qualitative and quantitive examples are shown in Fig.~\ref{intro2} and Tab.~\ref{tab:sam}. This significant improvement is attributed as follows. First, SAM sends the linguistic and visual features into the mask decoder directly, ignoring the multi-modal fusion. Second, since SAM is designed for interactive segmentation, features from its encoder are dominated by visual information. As a result, the subsequent mask decoder tends to segment the visually salient entity instead of the correct referring region. Third, SAM only uses global features from the last layer to get the precise location for the visual prompts, $i.e.$, points or bounding boxes. However, the text prompt also contains some expressions about grained details, which requires the model to focus on.

To adapt SAM into RIS, we propose a novel parameter-efficient tuning framework. First, we propose Mutual-Aware Attention to fuse the multi-modal features effectively. Second, we design the Mutual-Aware Mask Decoder to enable extra linguistic guidance. Last, we introduce the Feature Enhancement module to make our model focus on both global and local features. In summary, our framework reveals a practicable strategy for leveraging SAM into RIS, which may inspire the following RIS research.

\section{Conclusion}
This paper proposes a novel referring image segmentation method called RISAM, which effectively uses mutual-aware attention features and incorporates the powerful knowledge from SAM into RIS through the parameter-efficient fine-tuning framework. Our model contains three components: the Feature Enhancement module, the Mutual-Aware Attention block, and the Mutual-Aware Mask Decoder. To be specific, the Feature Enhancement module incorporates global and local features to transfer the knowledge from the frozen image encoder of SAM. Subsequently, the Mutual-Aware Attention block produces language-aware visual features and vision-aware linguistic features by weighting each word of the sentence and each region of visual features. Finally, we design the Mutual-Aware Mask Decoder to utilize mutual-aware attention features effectively by introducing an extra linguistic guidance. Besides, we introduce the multi-modal query token to integrate visual and linguistic properties, which is beneficial to a language-consistent mask. Extensive experiments on three well-known benchmarks and PhraseCut demonstrate that RISAM achieves new state-of-the-art performance and great generalization ability. Additionally, the performance on gRefCOCO verifies that our model is also competitive on multi-object referring image segmentation. 

\bibliographystyle{ACM-Reference-Format}
\bibliography{sample-base}

\end{document}